\documentclass[conference]{IEEEtran}
\IEEEoverridecommandlockouts
\usepackage{cite}
\usepackage{amsmath,amssymb,amsfonts}
\usepackage{algorithmic}
\usepackage{graphicx}
\usepackage{textcomp}
\usepackage{xcolor, colortbl}
\usepackage{footnote}
\usepackage{url}
\usepackage{subcaption}
\usepackage{authblk}
\def\BibTeX{{\rm B\kern-.05em{\sc i\kern-.025em b}\kern-.08em
    T\kern-.1667em\lower.7ex\hbox{E}\kern-.125emX}}
\begin{document}

\title{ML-Based Approach for NFL Defensive Pass Interference Prediction Using GPS Tracking Data}

\author[1]{Arian Skoki}
\author[1,2]{Jonatan Lerga}
\author[1,2]{Ivan Štajduhar}
\affil[1]{\textit{University of Rijeka, Faculty of Engineering, Department of Computer Engineering}, Rijeka 51000, Croatia \authorcr Email: {\tt \{askoki, jlerga, istajduh\}@riteh.hr}}
\affil[2]{\textit{University of Rijeka, Center for Artificial Intelligence and Cybersecurity}, Rijeka 51000, Croatia}


\maketitle

\begin{abstract}
Defensive Pass Interference (DPI) is one of the most impactful penalties in the NFL. DPI is a spot foul, yielding an automatic first down to the team in possession. With such an influence on the game, referees have no room for a mistake. It is also a very rare event, which happens 1-2 times per 100 pass attempts. With technology improving and many IoT wearables being put on the athletes to collect valuable data, there is a solid ground for applying machine learning (ML) techniques to improve every aspect of the game. The work presented here is the first attempt in predicting DPI using player tracking GPS data. The data we used was collected by NFL’s Next Gen Stats throughout the 2018 regular season. We present ML models for highly imbalanced time-series binary classification: LSTM, GRU, ANN, and Multivariate LSTM-FCN. Results showed that using GPS tracking data to predict DPI has limited success. The best performing models had high recall with low precision which resulted in the classification of many false positive examples. Looking closely at the data confirmed that there is just not enough information to determine whether a foul was committed. This study might serve as a filter for multi-step pipeline for video sequence classification which could be able to solve this problem. 
\end{abstract}


\section{Introduction}

American Football is the most popular sport in the United States. National Football League (NFL) is a professional American football league that consists of 32 teams, divided in two conferences and many divisions. Each team plays a regular part of the season and then, if successful, proceeds to the playoffs. On the field, there are 11 players in every team and the goal of the game is to win as many yards as possible and ultimately score more points than the other team. An American football field is 100 yards (91.44 m) long and 160 feet (48.8 m) wide. Points are awarded on touchdown (6 points), field goal (3 points), safety (2 points) and try after touchdown (1-2 points)~\footnote{\url{https://operations.nfl.com/the-rules/2020-nfl-rulebook/} (last accessed on 26. May 2021)}.

One of the biggest game-changers in NFL is the defensive pass interference (DPI) penalty which applies from the time the ball is thrown until the ball is touched. The penalty for DPI is an automatic first down at the spot of the foul. The official rule for pass interference is any act performed by a player more than one yard beyond the line of scrimmage which significantly hinders an eligible player’s opportunity to catch the ball. Pass interference can only occur when a forward pass is thrown from behind the line of scrimmage, regardless of whether the pass is legal or illegal, or whether it crosses the line~\footnote{\url{https://operations.nfl.com/the-rules/nfl-video-rulebook/defensive-pass-interference/} (last accessed on 26. May 2021)}.

NFL Data Bowl 2019 provided data of predicted catch probabilities. This also opened up an opportunity to analyse $\approx$6,000 catch plays from a sample of 91 games. Peter Wu and Brendon Gu found the bimodal shape of the defensive pass interference (DPI) distributions, which could be a result of differing standards for what a DPI entails \cite{DIRECT}. In their paper "DIRECT: A Two-Level System for Defensive Pass Interference Rooted in Repeatability, Enforceability, Clarity, and Transparency", they have presented a new approach for lowering the influence of DPI on the game~\cite{DIRECT}. In the proposed paper, only the obvious fouls, where the defender had no intention to play the ball, could be penalised in a spot foul.

In recent years, data analytics in NFL has become quite important, and more and more data is made publicly available for enthusiasts and scientists to explore. This year's Data Bowl (2021), among other questions, asked if there is any way player tracking data can be used to predict whether or not specific penalties - for example, defensive pass interference - will be called? This paper is, to the best of our knowledge, the first attempt in predicting DPI using tracking (GPS) data.

\section{Materials and Methods}

\subsection{Data Acquisition}

Data was acquired from "NFL Big Data Bowl 2021" which was hosted on \textit{Kaggle}~\footnote{\url{https://www.kaggle.com/c/nfl-big-data-bowl-2021/}}. The data was provided for competition, non-commercial and academic usage. The 2021 Big Data Bowl data contains player tracking, play, game, and player level information for all possible passing plays during the 2018 regular season. This means that 17 weeks of a typical NFL season contain only 259 actions that resulted in a DPI, which is 1.46\% of all actions (17,703). Such a data set is highly imbalanced, which can be compared with the problem of card fraud detection classification~\cite{card_frauds}. 

\subsection{Data Processing}\label{sec:data_processing}

The raw data set contains the information concerning the majority of players involved in that particular action play. Every play is divided into timestamp segments. Every timestamp contains tracking data of all defenders, attackers, and the ball. This can give a maximum of 23 records in a single timestamp. Keeping in mind the amount of data, that is simply too much information for a model to generalise on. A model would have to learn the change of patterns for each variable (distance, speed, acceleration, etc.) for every player and additionally, a relationship of these variables between the players. DPI occurs between players competing for the ball. Therefore, the focus in processing should be on extracting the most relevant players and ball information from every given play.

\begin{figure}[!tb]
  \begin{subfigure}[b]{1\columnwidth}
    \centering
     \includegraphics[width=1\columnwidth]{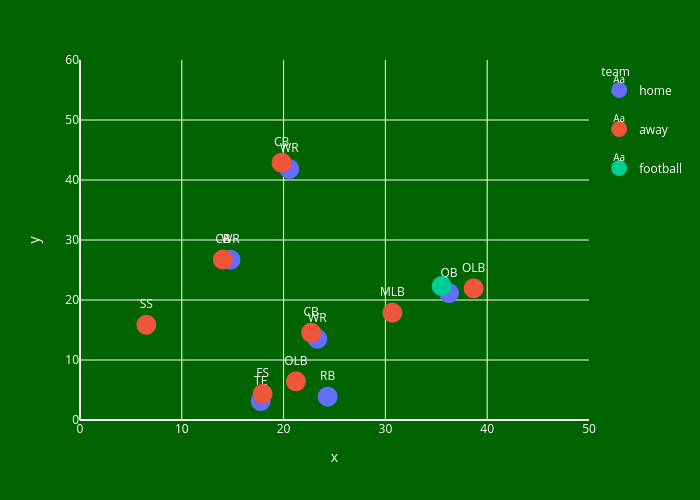}
     \subcaption{QB throws the ball to the WR.}
     \label{subfig:qb_pass}
  \end{subfigure}
  
  \begin{subfigure}[b]{1\columnwidth}
    \centering
     \includegraphics[width=1\columnwidth]{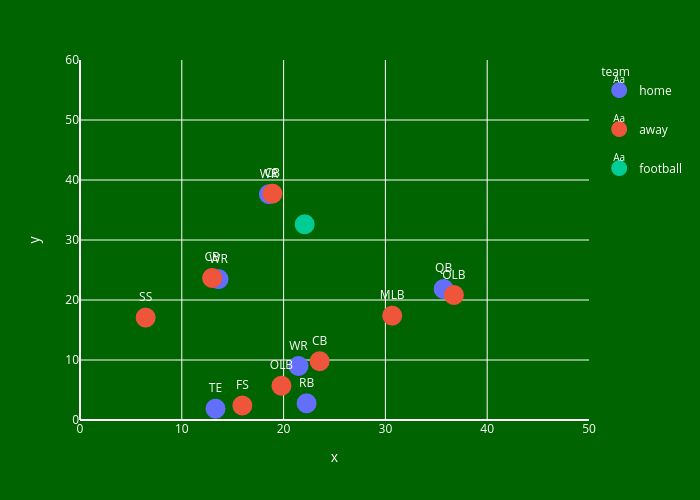}
     \subcaption{WR and CB fighting for the ball which is in the air.}
     \label{subfig:ball_air}
  \end{subfigure}
  
  \begin{subfigure}[b]{1\columnwidth}
    \centering
     \includegraphics[width=1\columnwidth]{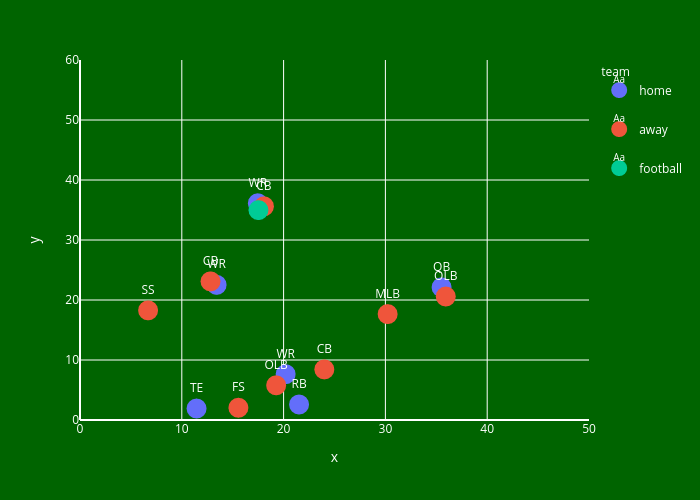}
     \subcaption{Ball arrives and DPI is committed.}
     \label{subfig:pass_arrived}
  \end{subfigure}
  
  \caption{Example of play which resulted in DPI}
  \label{fig:dpi_play}
\end{figure}

An example of DPI play is shown on image \ref{fig:dpi_play}. Blue and red circles represent the home and the away team respectively. The ball is depicted with a green circle, and the blue team has possession. The width of the pitch is represented on the y-axis. Height is subtracted from the original 120 yards to 50 yards, to provide better visualisation of this particular play. DPI can occur from the moment the ball is thrown forward by a quarterback (QB) and that moment is represented in image~\ref{subfig:qb_pass}. In the image~\ref{subfig:ball_air}, the ball is in the air and on its way to the target wide receiver (WR). Image~\ref{subfig:pass_arrived} shows the last moments of the play, where cornerback (CB) and WR are fighting for the ball and CB is committing a DPI.

\begin{figure}[!tb]
    \centering
    \includegraphics{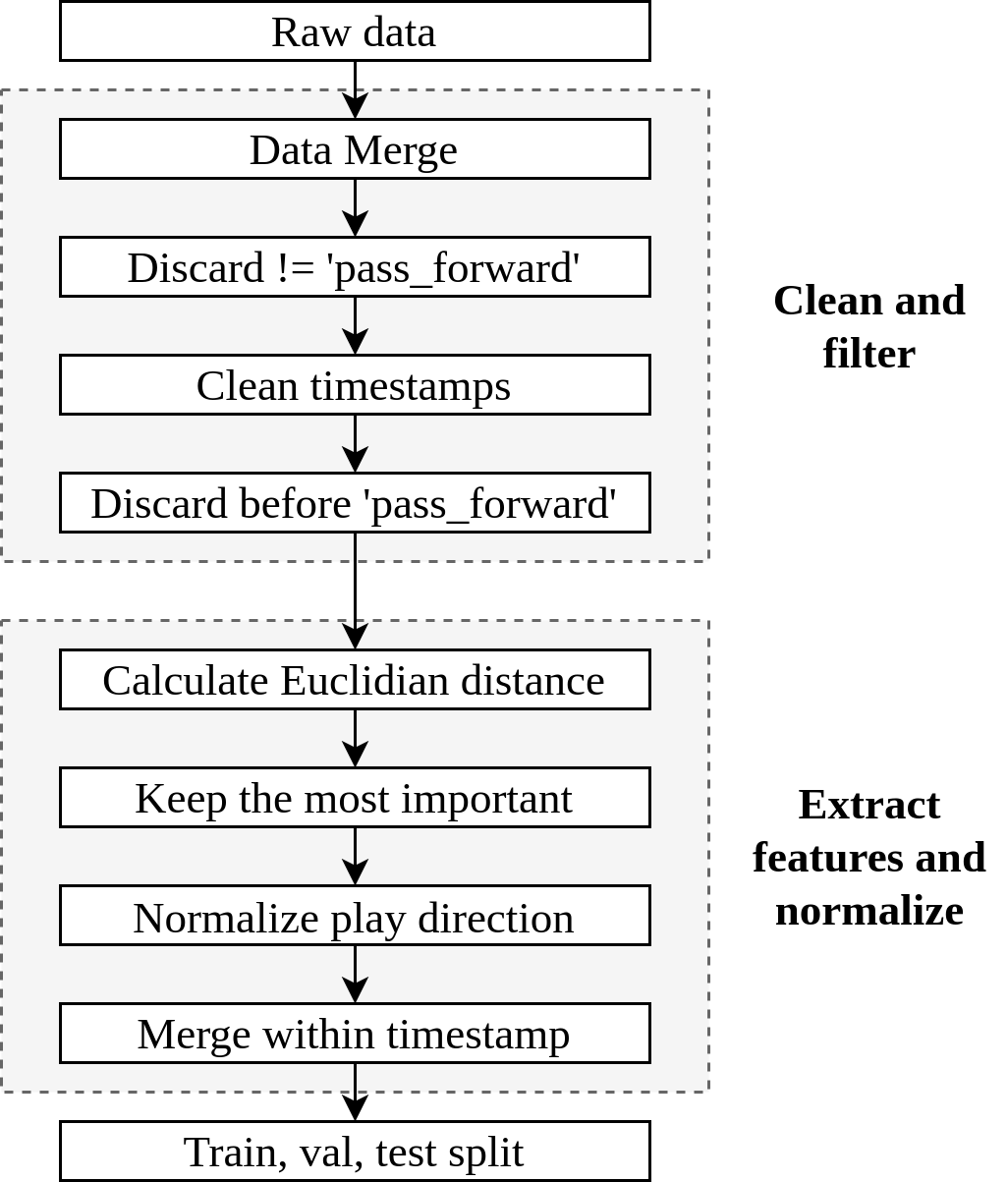}
    \caption{Preprocessing steps}
    \label{fig:preprocessing}
\end{figure}

Processing starts with merging plays, games and 17 weeks of tracking data. As DPI can only occur when a pass is played forward, every play that does not contain the "pass\_forward" event is discarded. Tracking data contains some duplicate timestamp values, therefore these had to be cleaned. All records were grouped according to game id, timestamp and frame id. Iterating through grouped data, records that had the same timestamp but different frame id were split by adding 10 ms on the previous timestamp. The minimal difference between timestamps is 100 ms, so this way, we have ensured that every timestamp within the play corresponds to the unique moment in the game. Next, all data which occurred before the "pass\_forward" event is discarded. Every play starts with the "pass\_forward" (image~\ref{subfig:qb_pass}) event and afterwards is a "None" event (image~\ref{subfig:ball_air}) which indicates that nothing important has happened. Next event, following these two, can be any: "tackle", "pass\_outcome\_caught", "pass\_unsuccessful", etc. That event, which follows, is interpreted as end-of-play, and that is the moment in which we choose the most important players responsible for the DPI. This situation is shown on image~\ref{subfig:pass_arrived}. The assumption is that, in a majority of cases, opposing players which are closest to the ball are the ones to watch for a DPI event. That is why Euclidean distance is calculated between the position of the ball and all the players. From each team, players having the smallest distance to the ball, in that particular moment, are picked (along with the ball), and a new data set is created. Additionally, data were normalised according to the play direction. That way, all the attacks lead in the same direction -- right. Also, it was important to distinguish the attacker from the defender, so that information was created from the available information concerning the team in possession. Attacker, defender and ball information were all merged in a single row, and that included acceleration, speed, orientation and direction information. Data that could not be merged easily was a current x and y position on the field. This was solved by calculating the Euclidean distance between all three parties: defender and attacker, attacker and the ball, defender and the ball. In addition to this data, most meaningful events were added as binary static variables. These events included: "pass\_arrived", "pass\_outcome\_caught", "tackle", "first\_contact", "pass\_outcome\_incomplete" and "out\_of\_bounds". All processing steps are shown in Fig.~\ref{fig:preprocessing}.

Players whose distance from each other (attacker and defender) is high, are unlikely to commit a foul as they are not close enough. In order to determine the threshold value for acceptance of a given play, DPI records were further examined. By examining the whole dataset, it proved that 90\% of DPI classified plays were below the value of 5.56 for the maximal distance between the defender and the attacker in a particular play. This threshold was then used on the whole dataset which resulted in the final distribution where the number of non-DPI plays was 9,529 (97.68\%) and the number of DPI plays was 231 (2.32\%). This filter eliminated possible outliers and enabled the focus to be shifted on the relevant samples. Everything was split into training (56\%), validation (14\%) and test (30\%) set with distributions shown in Table~\ref{table:data_distribution}. 

\begin{table}[!tb]
    \caption{Data set class distribution.}
    \begin{center}
        \begin{tabular}{|c|c|c|c|c|}
        \hline
        \textbf{} & \textbf{Training} & \textbf{Validation}& \textbf{Test} & \textbf{\%}\\
        \hline
        \textbf{non-DPI} & 5336 & 1334 & 2859 & 97.6 \\
        \hline
        \textbf{DPI} & 130 & 32 & 69 & 2.4 \\
        \hline
        \end{tabular}
    \end{center}
    \label{table:data_distribution}
\end{table}

\subsection{Prediction Models}

A sequence of events is important when considering DPI. That is why we opted for using time-series models for binary classification. The following include: Long-Short Term Memory (LSTM), Gated Recurrent Unit (GRU), Attention Neural Network (ANN) and, finally, multivariate LSTM Fully Convolutional Network (MLSTM-FCN)\cite{mlstm}. 

LSTM is a recurrent neural network (RNN), commonly used for tasks such as handwriting recognition, speech recognition and anomaly detection. An LSTM unit is composed of a cell, an input gate, an output gate and a forget gate. They are well suited for making predictions based on time series data and are able to cope with the vanishing gradient problem which can occur when using regular RNNs~\cite{lstm_time_series}.

The GRU is somewhat similar to an LSTM unit -- it has a forget gate, however, it contains fewer parameters, and has no output gate. It is used in polyphonic music modelling, speech signal modelling and natural language processing (NLP). The GRU was picked because it has been shown to exhibit better performance on smaller datasets~\cite{gru_performance}.

The attention mechanism in an ANN model focuses on the important parts of the data and fades out the rest. ANNs are mainly used in NLP and computer vision. It is an upgrade on the LSTM approach and it was extensively used in transformer networks~\cite{attention_transformer}.

MLSTM-FCN is an approach using a combination of convolutional layers and LSTM units \cite{mlstm}. It showed very good results on various, even unbalanced datasets. The authors have also provided their code implementation of MLSTM-FCN~\footnote{\url{https://github.com/titu1994/LSTM-FCN} (last accessed on 26. May 2021)}.

LSTM and GRU models were implemented using Keras deep learning API and Tensorflow 2.0 library. ANN model was constructed using Keras Attention Mechanism library~\footnote{\url{https://github.com/philipperemy/keras-attention-mechanism} (last accessed on 26. May 2021)} built on top of an LSTM model. All models consisted of only one hidden layer, having either 8, 64 or 128 neurons. Experimental testing proved that increasing the number of layers did not improve model performance. Furthermore, it significantly increased model training time, thus in this paper we focus only on models having only one layer and a variable number of hidden cells.

\subsection{Handling imbalanced data}

Dataset for predicting DPI is highly imbalanced which needs to be taken into consideration when training a model and interpreting the results. This issue is usually solved by using techniques such as oversampling the minority class, undersampling the majority class, creating new artificial data using algorithms such as SMOTE and by manipulating class weights~\cite{handling_imbalanced}. The process of undersampling was tested but it did not result in a satisfying performance outside the training set. Generating new artificial data is difficult, considering multiple variables changing over time, so this approach was not considered. Oversampling is very similar to changing class weights so we have focused on the latter. Several weight factors were tested and, in the end, class weight formula~\ref{eq:weight_factor} proved to work the best with class 0 (non-DPI) weights being 0.51 and class 1 (DPI) being 20.52. It is calculated using the following expression:

\begin{equation} \label{eq:weight_factor}
w_{class}=\frac{n\_inst}{n\_classes * n\_inst_{class}},
\end{equation}
where $w_{class}$ represents the calculated weight for a given class, $n\_inst$ denotes the number of all instances in a dataset, $n\_classes$ denotes the number of distinct classes, and $n\_inst_{class}$ denotes the number of instances of a given class.

\section{Results}\label{sec:results}

Having in mind the imbalanced nature of the problem, classification accuracy was not considered as an appropriate model evaluation metric. Instead, we used precision, recall, area under the curve (AUC) and F1 score. From the mentioned metrics, we marked recall to be the most important one because we want as few missed DPI classifications as possible.

All 12 model combinations have been trained 5 times each, and from these, only those exhibiting the best performance on the validation set were picked. Models tried to reproduce the best precision at recall threshold being 0.8. A recall is the most important metric as it would be better to predict a false DPI and then check it manually (using video replay of the action) than to miss it. Model performance on the test set can be seen in Table~\ref{table:model_results}. 

\begin{table}[!tb]
    \caption{Model performance on the test set.}
    \begin{center}
        \begin{tabular}{|c|c|c|c|c|c|}
        \hline
        \textbf{Model} & \textbf{Cells} & \textbf{Recall} & \textbf{Precision}& \textbf{F1} & \textbf{AUC}\\
        \hline
        LSTM & 8 & 0.826 & 0.08 & 0.147 & 0.796 \\
        \hline
        LSTM & 64 & 0.855 & \textbf{0.091} & \textbf{0.164} & \textbf{0.821} \\
        \hline
        \rowcolor{lightgray} LSTM & 128 & \textbf{0.884} & 0.0748 & 0.138 & 0.807 \\
        \hline
        ANN & 8 & 0.841 & 0.072 & 0.133 & 0.787 \\
        \hline
        ANN & 64 & \textbf{0.884} & 0.073 & 0.135 & 0.803 \\
        \hline
        ANN & 128 & 0.855 & 0.076 & 0.139 & 0.798 \\
        \hline
        GRU & 8 & 0.855 & 0.023 & 0.046 & 0.487 \\
        \hline
        GRU & 64 & 0.87 & 0.074 & 0.137 & 0.801 \\
        \hline
        GRU & 128 & \textbf{0.884} & 0.072 & 0.133 & 0.801 \\
        \hline
        LSTM-FCN & 8 & 0.551 & 0.079 & 0.137 & 0.695 \\
        \hline
        LSTM-FCN & 64 & 0.609 & 0.068 & 0.122 & 0.701 \\
        \hline
        LSTM-FCN & 128 & 0.855 & 0.05 & 0.095 & 0.727 \\
        \hline
        \end{tabular}
    \end{center}
    \label{table:model_results}
\end{table}

\begin{figure}[!tb]
    \centering
    \includegraphics[width=0.45\textwidth]{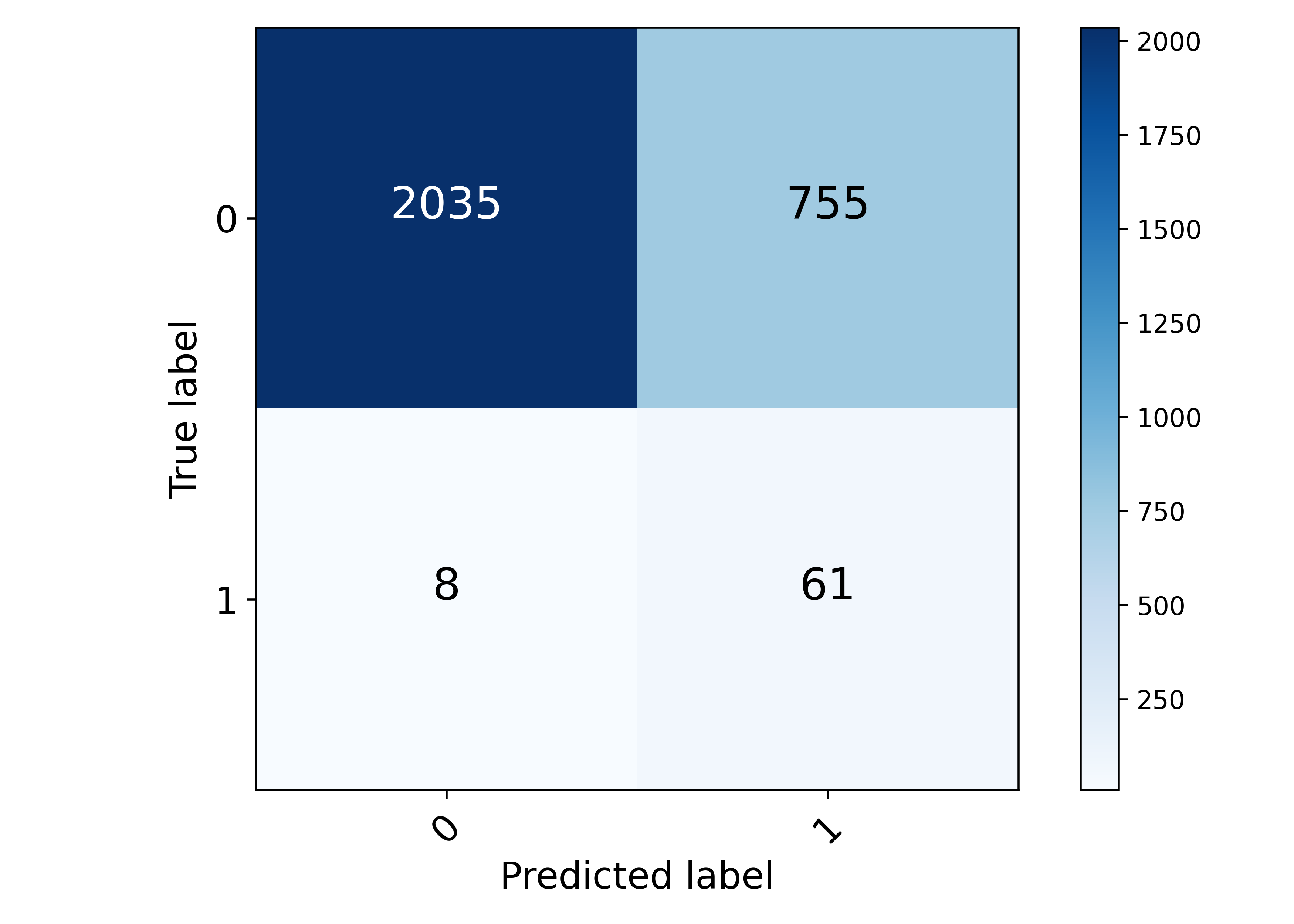}
    \caption{Classification confusion matrix for the LSTM (128) model on the test set.}
    \label{fig:cm_lstm_128}
\end{figure}

Best performing scores are emphasised. Three models achieved the best recall of 0.884 on the test set. In order to provide additional information about generalisation quality of models, both validation (V) and test (T) set performance is presented in Table~\ref{table:model_results_compare}. The best model according to all other metrics (precision, F1 and AUC) was the LSTM model having 64 hidden-layer neurons. Out of the models exhibiting the highest recall, the best performing model considering precision is the LSTM model having 128 hidden-layer neurons. Classification confusion matrix for this model is shown in Fig.~\ref{fig:cm_lstm_128}. Model performance is discussed in more detail in section~\ref{sec:discussion}. In addition to the presented results, we provide full code for preprocessing, training and evaluating models that will simplify future research in this area and enable easier replication of results~\footnote{The code is available at \url{https://github.com/askoki/nfl_dpi_prediction}}.

\begin{table}[!tb]
    \caption{Model performance on test (T) versus validation (V) set}
    \begin{center}
        \begin{tabular}{|c|c|c|c|c|}
        \hline
        \textbf{Model} & \textbf{Recall(T)} & \textbf{Recall(V)} & \textbf{Precision(T)} & \textbf{Precision(V)}\\
        \hline
        LSTM (64) & \textbf{0.855} & 0.844 & \textbf{0.091} & 0.085 \\
        \hline
        LSTM (128) & 0.884 & \textbf{0.938} & 0.075 & \textbf{0.078} \\
        \hline
        ANN (64) & 0.884 & \textbf{0.969} & 0.073 & \textbf{0.079} \\
        \hline
        GRU (128) & 0.884 & \textbf{0.938} & 0.072 & \textbf{0.077} \\
        \hline
        \end{tabular}
    \end{center}
    \label{table:model_results_compare}
\end{table}

\section{Discussion}\label{sec:discussion}

The models presented in this work fail to solve the problem of detecting seldom occurring events such as DPI. Changing weight factors, increasing model complexity or changing the sequence model algorithm all fail to deliver better results, and an F1 score significantly above 0.15. By looking at the source of the data -- GPS sensors -- this comes as no surprise. When players are close to each other, there is no information from which one can determine if a DPI was made. Our initial thought was that players' trajectories might obfuscate the information concerning a possible DPI, but this was just not enough to detect this event with high prediction confidence. Initially, all players (on the pitch) were involved as part of the information from which the models were trained. This, however, did not give any meaningful results -- everything was classified as non-DPI (the dominant class). We then switched our focus to include only those players having the highest probability to commit a DPI, which was described in section~\ref{sec:data_processing}. Changes in data processing resulted in the improvement of model performance but all of them failed to deliver a recall greater than 0.88 as presented in section \ref{sec:results}.

An additional challenge when dealing with this problem was the high data imbalance. Furthermore, there were only 259 possible plays that resulted in a DPI. This amount of data might not be enough for the ML models to build on~\cite{imbalanced_dataset}. Gathering DPI plays from more seasons could be beneficial in improving model efficiency, at least in recall metrics. Improving recall, while keeping a satisfying precision score, would be a useful automatic preprocessing step for a follow-up video analysis filter.

Images capture exact events on the pitch, therefore models could learn if a player made an illegal act or not~\cite{video_pattern_recognition}. On the other hand, tracking data does not provide any additional information when players are close to each other and competing for the ball. Analysing sequences of images would provide more features for a model to work with, which could consequently lead to better classification performance. Again, the number of DPI data instances would have to be greater, but this approach does not suffer from a lack of information. European football (soccer) is using video recordings for a better understanding of the game~\cite{soccer_video}. In future work, combining model results from GPS and video systems might lead to a solution to this highly-complex problem~\cite{combine_gps_video}.

\section{Conclusion}

The work presented in this paper is the first approach to predict DPI events in American football using GPS tracking data. DPI is a very rare event occurring in only 1.46\% of all action plays. However, the impact of this penalty call can be game-changing. By automating this penalty call, the possibility of one referee call being a game changer would be drastically decreased. The results show that the prediction of this event, using ML sequence models, has limited applicability. The models failed to achieve a recall greater than 0.884 with precision being very low, usually around 0.08. 

The data set used for training contained only 259 DPI events. By increasing the sample size, the recall score could be potentially improved and this approach could be used as a first step filter for later video sequence analysis of remaining DPI candidates. On the given dataset, GPS tracking data alone does not contain enough information in order to classify this complex event correctly. Future work should take into consideration the amount of DPI plays which is available and try different approaches, such as video analysis, in order to try to improve model performance.

\section*{Acknowledgment}
This work was supported by the Horizon 2020 project EuroCC 951732 \textit{National Competence Centres in the Framework of EuroHPC}, and by the University of Rijeka, Croatia [grant number uniri-tehnic-18-15 and uniri-tehnic-18-17].

\bibliographystyle{IEEEtran}
\bibliography{nfl}

\begin{thebibliography}{10}
\providecommand{\url}[1]{#1}
\csname url@samestyle\endcsname
\providecommand{\newblock}{\relax}
\providecommand{\bibinfo}[2]{#2}
\providecommand{\BIBentrySTDinterwordspacing}{\spaceskip=0pt\relax}
\providecommand{\BIBentryALTinterwordstretchfactor}{4}
\providecommand{\BIBentryALTinterwordspacing}{\spaceskip=\fontdimen2\font plus
\BIBentryALTinterwordstretchfactor\fontdimen3\font minus
  \fontdimen4\font\relax}
\providecommand{\BIBforeignlanguage}[2]{{%
\expandafter\ifx\csname l@#1\endcsname\relax
\typeout{** WARNING: IEEEtran.bst: No hyphenation pattern has been}%
\typeout{** loaded for the language `#1'. Using the pattern for}%
\typeout{** the default language instead.}%
\else
\language=\csname l@#1\endcsname
\fi
#2}}
\providecommand{\BIBdecl}{\relax}
\BIBdecl

\bibitem{DIRECT}
P.~Wu and B.~Gu, ``\BIBforeignlanguage{en}{{DIRECT}: {A} {Two}-{Level} {System}
  for {Defensive} {Pass} {Interference} {Rooted} in {Repeatability},
  {Enforceability}, {Clarity}, and {Transparency}},'' p.~12.

\bibitem{card_frauds}
S.~Mittal and S.~Tyagi, ``Performance {Evaluation} of {Machine} {Learning}
  {Algorithms} for {Credit} {Card} {Fraud} {Detection},'' in \emph{2019 9th
  {International} {Conference} on {Cloud} {Computing}, {Data} {Science}
  {Engineering} ({Confluence})}, Jan. 2019, pp. 320--324.

\bibitem{mlstm}
\BIBentryALTinterwordspacing
F.~Karim, S.~Majumdar, H.~Darabi, and S.~Harford, ``Multivariate {LSTM}-{FCNs}
  for {Time} {Series} {Classification},'' \emph{Neural Networks}, vol. 116, pp.
  237--245, Aug. 2019, arXiv: 1801.04503. [Online]. Available:
  \url{http://arxiv.org/abs/1801.04503}
\BIBentrySTDinterwordspacing

\bibitem{lstm_time_series}
S.~Siami-Namini, N.~Tavakoli, and A.~S. Namin, ``The {Performance} of {LSTM}
  and {BiLSTM} in {Forecasting} {Time} {Series},'' in \emph{2019 {IEEE}
  {International} {Conference} on {Big} {Data} ({Big} {Data})}, Dec. 2019, pp.
  3285--3292.

\bibitem{gru_performance}
S.~Yang, X.~Yu, and Y.~Zhou, ``{LSTM} and {GRU} {Neural} {Network}
  {Performance} {Comparison} {Study}: {Taking} {Yelp} {Review} {Dataset} as an
  {Example},'' in \emph{2020 {International} {Workshop} on {Electronic}
  {Communication} and {Artificial} {Intelligence} ({IWECAI})}, Jun. 2020, pp.
  98--101.

\bibitem{attention_transformer}
\BIBentryALTinterwordspacing
N.~Li, S.~Liu, Y.~Liu, S.~Zhao, and M.~Liu, ``Neural {Speech} {Synthesis} with
  {Transformer} {Network},'' \emph{Proceedings of the AAAI Conference on
  Artificial Intelligence}, vol.~33, pp. 6706--6713, Jul. 2019. [Online].
  Available: \url{https://aaai.org/ojs/index.php/AAAI/article/view/4642}
\BIBentrySTDinterwordspacing

\bibitem{handling_imbalanced}
\BIBentryALTinterwordspacing
E.-A. Mînăstireanu and G.~Meșniță, ``\BIBforeignlanguage{en}{Methods of
  {Handling} {Unbalanced} {Datasets} in {Credit} {Card} {Fraud} {Detection}},''
  \emph{\BIBforeignlanguage{en}{BRAIN. Broad Research in Artificial
  Intelligence and Neuroscience}}, vol.~11, no.~1, pp. 131--143, Mar. 2020,
  number: 1. [Online]. Available:
  \url{https://www.edusoft.ro/brain/index.php/brain/article/view/994}
\BIBentrySTDinterwordspacing

\bibitem{imbalanced_dataset}
\BIBentryALTinterwordspacing
M.~Maalouf and T.~B. Trafalis, ``\BIBforeignlanguage{en}{Rare events and
  imbalanced datasets: an overview},''
  \emph{\BIBforeignlanguage{en}{International Journal of Data Mining, Modelling
  and Management}}, vol.~3, no.~4, p. 375, 2011. [Online]. Available:
  \url{http://www.inderscience.com/link.php?id=42935}
\BIBentrySTDinterwordspacing

\bibitem{video_pattern_recognition}
R.~Chellappa, A.~Veeraraghavan, and G.~Aggarwal,
  ``\BIBforeignlanguage{en}{Pattern {Recognition} in {Video}},'' in
  \emph{\BIBforeignlanguage{en}{Pattern {Recognition} and {Machine}
  {Intelligence}}}, ser. Lecture {Notes} in {Computer} {Science}, S.~K. Pal,
  S.~Bandyopadhyay, and S.~Biswas, Eds.\hskip 1em plus 0.5em minus 0.4em\relax
  Berlin, Heidelberg: Springer, 2005, pp. 11--20.

\bibitem{soccer_video}
M.~Stein, H.~Janetzko, A.~Lamprecht, T.~Breitkreutz, P.~Zimmermann,
  B.~Goldlücke, T.~Schreck, G.~Andrienko, M.~Grossniklaus, and D.~A. Keim,
  ``Bring {It} to the {Pitch}: {Combining} {Video} and {Movement} {Data} to
  {Enhance} {Team} {Sport} {Analysis},'' \emph{IEEE Transactions on
  Visualization and Computer Graphics}, vol.~24, no.~1, pp. 13--22, Jan. 2018,
  conference Name: IEEE Transactions on Visualization and Computer Graphics.

\bibitem{combine_gps_video}
\BIBentryALTinterwordspacing
``\BIBforeignlanguage{en}{Combining {Video} and {Player} {Telemetry} for
  {Evidence}-based {Decisions} in {Soccer}:},'' in
  \emph{\BIBforeignlanguage{en}{Proceedings of the {International} {Congress}
  on {Sports} {Science} {Research} and {Technology} {Support}}}.\hskip 1em plus
  0.5em minus 0.4em\relax Vilamoura, Algarve, Portugal: SCITEPRESS - Science
  and and Technology Publications, 2013, pp. 197--205. [Online]. Available:
  \url{http://www.scitepress.org/DigitalLibrary/Link.aspx?doi=10.5220/0004676101970205}
\BIBentrySTDinterwordspacing

\end{thebibliography}

\vspace{12pt}
\color{red}

\end{document}